\pdfoutput=1

\documentclass[11pt]{article}

\usepackage{ACL2023}

\usepackage{times}
\usepackage{latexsym}
\usepackage{graphicx}
\usepackage{amssymb}
\usepackage{amsmath}
\usepackage{amsfonts}
\usepackage[ruled,vlined]{algorithm2e}
\usepackage{booktabs}
\usepackage{multirow}
\usepackage{subcaption}
\usepackage{marvosym}
\usepackage{xspace}

\newcommand{\modelname}[0]{{\textsc{Seer}}\xspace}

\usepackage[T1]{fontenc}

\usepackage[utf8]{inputenc}

\usepackage{microtype}

\usepackage{inconsolata}

\title{\textsc{Seer}: Facilitating Structured Reasoning and Explanation via Reinforcement Learning}

\author{Guoxin Chen\textsuperscript{\dag\S}, Kexin Tang\textsuperscript{$\ast$}, Chao Yang\textsuperscript{\dag\Letter}, Fuying Ye, Yu Qiao\textsuperscript{\dag}, Yiming Qian\textsuperscript{\ddag\Letter}\\
\textsuperscript{\dag}Shanghai Artificial Intelligence Laboratory \\
\textsuperscript{\S}Institute of Computing Technology, Chinese Academy of Sciences \\
\textsuperscript{$\ast$}Shanghai Jiao Tong University \\
\textsuperscript{\ddag}Agency for Science, Technology and Research (A*STAR) \\
\texttt{chenguoxin22s@ict.ac.cn,} 
\texttt{\{tkx94china, fuyingye.work\}@gmail.com,} \\
\texttt{\{yangchao,qiaoyu\}@pjlab.org.cn,}
\texttt{qiany@ihpc.a-star.edu.sg}
}

\begin{document}
\maketitle
\begin{abstract}
Elucidating the reasoning process with structured explanations from question to answer is crucial, as it significantly enhances the interpretability, traceability, and trustworthiness of question-answering (QA) systems.
However, structured explanations demand models to perform intricately structured reasoning, which poses great challenges.
Most existing methods focus on single-step reasoning through supervised learning, ignoring logical dependencies between steps.
Moreover, existing reinforcement learning (RL) based methods overlook the structured relationships, underutilizing the potential of RL in structured reasoning.
In this paper, we propose \modelname, a novel method that maximizes a structure-based return to facilitate structured reasoning and explanation.
Our proposed structure-based return precisely describes the hierarchical and branching structure inherent in structured reasoning, effectively capturing the intricate relationships between different reasoning steps.
In addition, we introduce a fine-grained reward function to meticulously delineate diverse reasoning steps.
Extensive experiments show that \modelname significantly outperforms state-of-the-art methods, achieving an absolute improvement of 6.9\% over RL-based methods on EntailmentBank, a 4.4\% average improvement on STREET benchmark, and exhibiting outstanding efficiency and cross-dataset generalization performance. Our code is available at \url{https://github.com/Chen-GX/SEER}.
\end{abstract}

\section{Introduction}
\begin{figure}[t]
    \centering
    \includegraphics[width=0.99\linewidth]{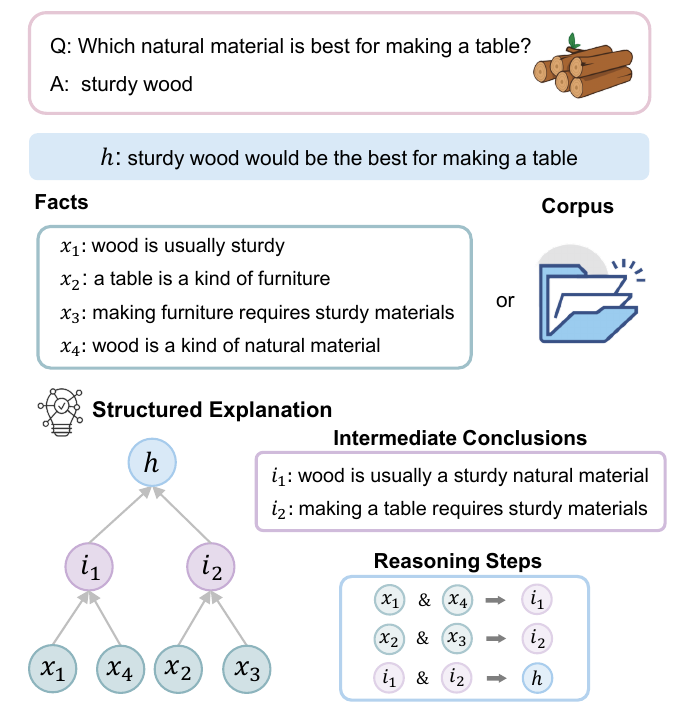}
    \caption{An example of structured explanation. Given a hypothesis $h$ (a declarative sentence derived from a question-answer pair) and a set of facts (or corpus), the goal is to generate a structured explanation, which delineates the reasoning process from facts to the hypothesis.}
    \label{fig:example_of_entailment_tree}
\end{figure}

Navigating machines to understand and articulate the thought process from posing a question to arriving at an answer has been a long-term pursuit in the AI community~\citep{mccarthy1959programs,yu2023nature}.
Current QA explainable systems adeptly furnish brief supporting evidence~\citep{rajani-etal-2019-explain,deyoung-etal-2020-eraser}.
However, they often fail to clarify the \textit{reasoning process} from prior knowledge to the derived answer.
By elucidating the reasoning process of answers generation from the language models, we can greatly improve interpretability, trustworthiness, and debuggability~\citep{dalvi-etal-2021-explaining,DBLP:conf/iclr/Ribeiro0MZDKBRH23}.
As illustrated in Figure~\ref{fig:example_of_entailment_tree}, when generating answers for the question "Which natural material is best for making a table?", the reasoning process with structured explanations, such as entailment trees~\citep{dalvi-etal-2021-explaining} or reasoning graphs~\citep{DBLP:conf/iclr/Ribeiro0MZDKBRH23}, explains why "sturdy wood" is the best answer.


\begin{table}[t]
\centering
    \resizebox{0.93\linewidth}{!}{
\begin{tabular}{@{}lccc@{}}
\toprule
Method & Training Emphasis        & Runtime & Return     \\ \midrule
RLET   & multi-step reasoning  & 9.34s   & chained    \\
FAME   & single-step reasoning & 30.77s  & /          \\
Ours   & structured reasoning  & 3.91s   & structured \\ \bottomrule
\end{tabular}
}
\caption{
Comparative analysis of different methods: RL-based method, RLET~\citep{liu-etal-2022-rlet}, supervised method, FAME~\citep{hong-etal-2023-faithful}, and our approach.
}
\label{tab:method_comparison}
\end{table}

Deriving such complex structured explanations poses a great challenge.
Previous methods~\citep{dalvi-etal-2021-explaining,tafjord-etal-2021-proofwriter} consider structured explanations as linearized sequences and generate the entire reasoning process in one go.
However, these methods lack controllability and may hallucinate unreliable reasoning steps.
To address these concerns, recent studies~\citep{hong-etal-2022-metgen,neves-ribeiro-etal-2022-entailment,yang-etal-2022-generating} decompose structured explanations and focus on single-step reasoning via supervised learning. Nevertheless, this kind of approach may not always yield optimal results as they fail to consider the interdependencies between different steps.
FAME~\citep{hong-etal-2023-faithful} attempts to compensate for these shortcomings by leveraging Monte-Carlo planning~\citep{kocsis2006bandit}, which significantly increases the running time and inadvertently explores numerous ineffective steps (as shown in Table~\ref{tab:method_comparison}).
Furthermore, FAME still concentrates on isolated single-step reasoning, which lacks support for structured reasoning. As a general framework for solving sequential decision-making problems, reinforcement learning (RL) is employed in RLET~\citep{liu-etal-2022-rlet} to enhance multi-step reasoning.
However, RLET defines the return (a.k.a.  cumulative reward) using the standard chain structure, thus lacking the ability to represent the tree~\citep{dalvi-etal-2021-explaining} or graph~\citep{DBLP:conf/iclr/Ribeiro0MZDKBRH23} logical structures inherent in structured reasoning. As a result, the potential of RL for structured reasoning is not fully exploited.

To address the above issues, we propose \modelname, a novel method that \textit{facilitates \textbf{S}tructured r\textbf{E}asoning and \textbf{E}xplanation via \textbf{R}einforcement learning}.
In structured reasoning, we observe that the logical dependencies between different steps no longer follow a chained trajectory but instead adhere to the inherent tree or graph structure.
Therefore, we propose the structure-based return to precisely describe a tree or graph logical structure, effectively capturing the complex interdependencies between different steps.
Additionally, we refine the reward function to meticulously delineate diverse reasoning steps, specifically targeting redundant ones that do not contribute to the final structured explanations.
Through experiments in Sec.~\ref{sec:param_sensitive}, we find that redundant steps represent the exploration in the environment, and appropriate penalization contributes to improved reasoning performance.


\noindent \textbf{Our contributions} are summarized as follows:

\noindent $\bullet$ We propose \modelname, a novel RL-based method that facilitates structured reasoning and explanation.
To our knowledge, \modelname is the first general framework that accommodates scenarios of chained, tree-based, and graph-based structured reasoning.


\noindent $\bullet$ We propose the structure-based return to address the intricate interdependencies among different reasoning steps, effectively stimulating the potential of RL in structured reasoning.

\noindent $\bullet$ We conduct extensive experiments to demonstrate the superiority of  \modelname over state-of-the-art methods. Our method facilitates the effectiveness and efficiency of structured reasoning and exhibits outstanding cross-dataset generalization performance.


\section{Related Work}
\subsection{Explanation for Question Answering}
Extensive research has delved into various forms of interpretability in QA systems~\citep{DBLP:journals/corr/abs-2010-00389,S2021_698d51a1,lamm2021qed,chen-etal-2023-mprompt}.
Different from the free-form texts susceptible to hallucinations~\citep{rajani-etal-2019-explain,NEURIPS2022_9d560961} or the rationales that only provide supporting evidence~\citep{deyoung-etal-2020-eraser,valentino-etal-2021-unification}, the structured explanations, such as the entailment trees~\citep{dalvi-etal-2021-explaining} and reasoning graphs~\citep{DBLP:conf/iclr/Ribeiro0MZDKBRH23}, 
offer a novel way to generate explanations. These structured methods utilize tree or graph formats to clearly outline \textit{what} information is used and \textit{how} it is combined to reach the answer.
Despite the remarkable interpretability, the intricately structured reasoning also poses significant challenges~\citep{yu2023nature,xu2023large}.

\subsection{Natural Language Reasoning}
Natural language reasoning, a process that integrates multiple knowledge to derive new conclusions, has attracted significant attention~\citep{saha-etal-2020-prover,tafjord-etal-2021-proofwriter,sanyal-etal-2022-fairr,chen2024alphamath}.
Among these, the entailment trees and reasoning graphs, which involve structured reasoning and reasoning path generation tasks, present considerable challenges~\citep{yu2023nature}.
\citet{dalvi-etal-2021-explaining} attempt to transform structured reasoning into a linearized sequence to fit generative models, which may generate hallucinations and invalid reasoning.
To alleviate this issue, recent studies~\citep{neves-ribeiro-etal-2022-entailment,hong-etal-2022-metgen,neves-ribeiro-etal-2022-entailment,hong-etal-2023-faithful} perform premises selection and reasoning in a step-by-step manner.
Nevertheless, these methods decompose structured reasoning and solely leverage isolated single-step supervision to train models. This kind of approach neglects the interdependencies between different steps, which may not always yield optimal results.
Therefore, in light of the advancements of RL in various reasoning tasks~\citep{NEURIPS2021_859555c7,NEURIPS2022_8636419d}, RLET~\citep{liu-etal-2022-rlet} attempts to incorporate RL into the entailment trees.
However, it has to enumerate all potential actions, which is unacceptable for practical scenarios. 
Furthermore, RLET still defines returns in chained trajectories to facilitate multi-step reasoning, which is not suitable for tree/graph-based structured reasoning.
In contrast, our \modelname showcases superior adaptability to chained, tree-based, and graph-based structured reasoning via the structure-based return, which significantly enhances both the reasoning performance and efficiency.

\section{Method}
\subsection{Task Formulation}
As illustrated in Figure~\ref{fig:example_of_entailment_tree}, the input of the task comprises a set of facts $X=\{x_1, x_2, \ldots, x_n\}$ and a hypothesis $h$. 
The output of the task is the reasoning steps in a structured form, such as an entailment tree $T$ or a reasoning graph\footnote{Although the reasoning graph~\citep{DBLP:conf/iclr/Ribeiro0MZDKBRH23} is a more general structure, to be consistent with the majority of previous work, we use the entailment tree~\citep{dalvi-etal-2021-explaining} as an example to formalize the task and illustrate our method. Our proposed method is also applicable to the task described in the form of a reasoning graph.}.
The entailment tree $T$ consists of tree-structured reasoning, whose leaf nodes are selected from the relevant facts ($x_{*}$) and intermediate nodes represent the derived intermediate conclusions ($i_{*}$).
We represent the annotated ground-truth entailment tree as $T_{gold}$, with its leaf nodes signifying $X_{gold}$. 
\begin{figure*}[ht]
    \centering
    \includegraphics[width=\linewidth]{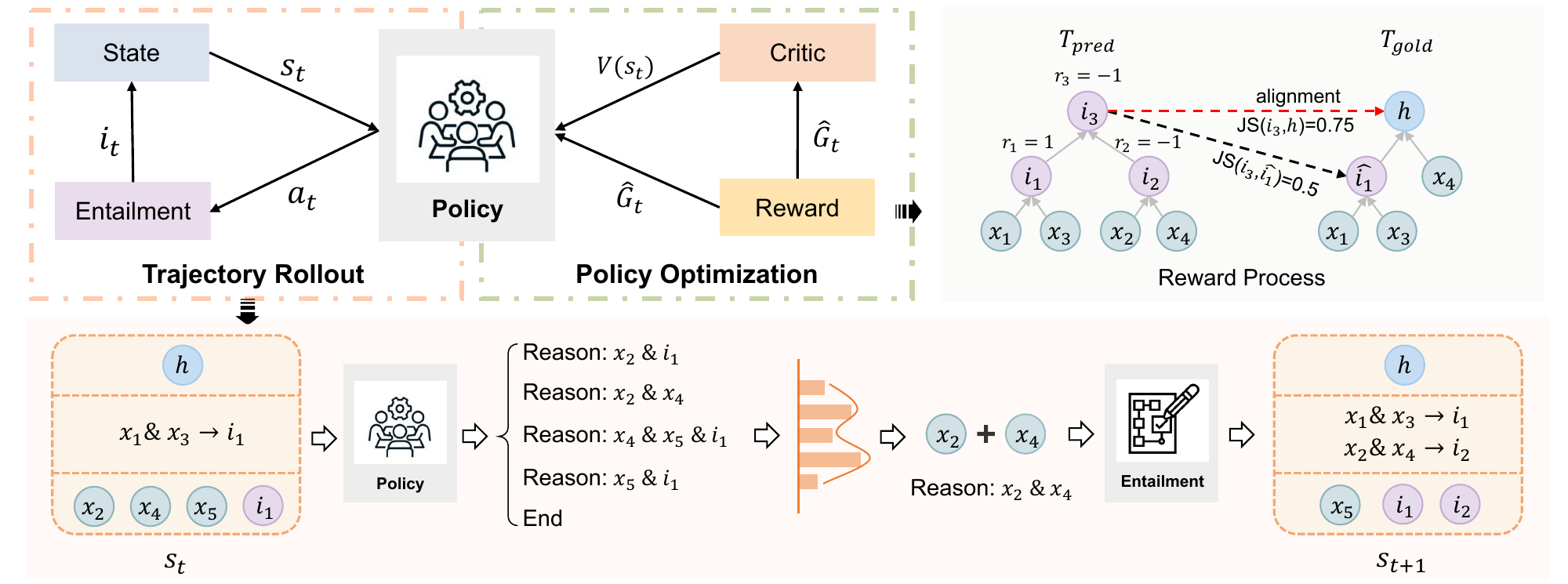}
    \caption{Overall framework of \modelname. For trajectory rollout, action generation (Policy) and conclusion generation (Entailment) are performed alternately. The orange area details the reasoning process from $s_t$ to $s_{t+1}$. For policy optimization, the reward module assigns rewards and updates the policy and critic based on tree or graph structures.}
    \label{fig:framework}
\end{figure*}
\subsection{Overview}
We model the structured reasoning as a reinforcement learning (RL) task, the goal of which is to learn the optimal reasoning policy.
Figure~\ref{fig:framework} illustrates the overall framework of \modelname, which mainly includes trajectory rollout and policy optimization.
For trajectory rollout, we generate trajectories based on the current policy, and each trajectory is produced iteratively until the stopping criteria are satisfied (Appendix~\ref{sec:appendix_stop_condition}).
For policy optimization, we assign rewards to the collected trajectories and update both the policy and critic using the structure-based return.
Algorithm~\ref{alg:train} (Appendix~\ref{sec:appendix_algorithm_details}) outlines our proposed method for further reference.


\subsection{Fine-grained Component of \modelname}

\paragraph{State}
At reasoning step $t$, we define the state $s_t = \{h, P_t, C_t\}$ as a combination of the hypothesis $h$, existing reasoning steps $P_t$ and candidate sentences $C_t$.
$P_t$ contains the reasoning steps so far, and $C_t$ is the set of sentences that can be selected as premises.
Each sentence in $C_t$ is either unused facts or intermediate conclusions $I_t$ generated by previous steps, i.e., $C_t = \{X\cup I_t \backslash U_t\}$, where $U_t$ is the set of used sentences.
For the initial state, $s_1 = \{h, P_1 = \varnothing, C_1=X\}$.

\paragraph{Action}
Given the state $s_t$, we consider two types of actions $a_t\in \mathcal{A}(s_t)$:
(1) "\texttt{Reason}: <premises>": the entailment module is invoked to generate a new intermediate conclusion $i_t$ based on the given <premises>.
Here, <premises> are selected from $C_t$.
Then, the state is updated as follows: $P_{t+1} = P_{t} \cup \{\text{\textless premises\textgreater} \rightarrow i_t\}$, $U_{t+1} = U_t\cup \{\text{\textless premises\textgreater}\}$, and $I_{t+1} = I_t \cup \{i_t\}$. (2) "\texttt{End}": This action signifies the end of the reasoning process and returns the trajectory $\tau$.

\paragraph{Policy}
The action type "\texttt{Reason}: <premises>" induces a large action space, since premises can be any combination of sentences from the candidate set $C$. To enumerate the probabilities of all potential actions and then sample an action to execute, previous studies~\citep{liu-etal-2022-rlet,hong-etal-2022-metgen} limit combinations to pairwise premises, such that the action space is reduced to $\binom{n}{2}$, where $n$ is the size of the set $C$. However, such a simplification incurs some potential drawbacks. First, as the number of candidate sentences increases, the number of potential actions grows exponentially. This renders them impractical for complex reasoning tasks with limited computational resources.
Second, by restricting combinations to pairs only, the interdependencies among multiple premises are ignored, which may limit the effectiveness and richness of the derived conclusions.


To address this issue, we adopt a generative model to represent the policy $\pi$, which can directly sample from the action space $\mathcal{A}(s_t)$. Using the generative model essentially expands the action space where the combinations of premises can be arbitrary. This enables the policy to extensively explore better actions during RL training, not limited to paired premises. Further, to speed up RL training, we first generate the top-$k$ actions using policy $\pi$:
\begin{equation}
    a_t^1, a_t^2, ..., a_t^k \sim \pi(a|s_t), \quad a \in \mathcal{A}(s_t),
\end{equation}
where the input is a linearized state $s_t$ (i.e., the concatenation of $h$, $P_t$, and $C_t$).
Then, we proceed with re-normalization to form an appropriate probability distribution over the top-$k$ actions, and sample from it to select the action $a_t$ to be performed in the current reasoning step, that is, 
\begin{equation}
    \pi'(a_t^i|s_t) = \frac{\pi(a_t^i|s_t)}{\sum_{j=1}^{k}\pi(a_t^j|s_t)}, \quad i=1,...,k,
    \label{eq:normalized_prob}
\end{equation}
\begin{equation}
    a_t \sim \pi'(a|s_t), \quad a \in \{a_t^1, a_t^2, ..., a_t^k\}.
\end{equation}

\paragraph{Entailment Module}
If the action $a_t$ is "\texttt{Reason}: <premises>", we invoke the entailment module to derive the intermediate conclusion to obtain the next state.
The entailment module is also a generative model with its input being <premises>. Following~\citet{hong-etal-2022-metgen,liu-etal-2022-rlet}, we fine-tune the entailment model in a supervised manner and freeze the parameters during the reinforcement learning process, as shown in Figure~\ref{fig:framework}.

\paragraph{Reward}
To evaluate the correctness of the entailment tree, \citet{dalvi-etal-2021-explaining} proposed an alignment algorithm based on Jaccard similarity to align each intermediate node of the predicted tree $T_{\text{pred}}$ with $T_{\text{gold}}$.
However, different from the fully supervised learning methods, we observe that during the RL process, the policy explores different actions to identify the optimal reasoning process, inevitably attempting some redundant steps that do not contribute to reaching the final hypothesis.
Existing RL-based work~\citep{liu-etal-2022-rlet} simply treats redundant steps with the same penalty as erroneous steps.
This simplification may negatively affect the learning process which discourages necessary exploration in the action space.
Furthermore, it lacks detailed feedback to guide the policy toward optimal policy, as it fails to differentiate between innocuous actions (redundant steps) and incorrect actions (erroneous steps).

To this end, we propose a fine-grained reward function that assigns different reward values for correct steps, erroneous steps, and redundant steps, as shown in Equation~\ref{eq:reward}.
For a trajectory $\tau$, we assume that the last intermediate conclusion is our predicted hypothesis since the policy deems it should \texttt{End} here.
Then, we backtrack to construct the predicted entailment tree $T_{\text{pred}}$ (see Appendix~\ref{sec:appendix_tree_construction} for more details).
Note that there might be some steps not participating in $T_{\text{pred}}$, which are regarded as redundant steps.
Then, as illustrated in Figure~\ref{fig:reward_process}, we consider steps that perfectly match via the alignment algorithm~\citep{dalvi-etal-2021-explaining} as correct steps and regard others as erroneous steps.
\begin{equation}
    r_t = 
\begin{cases} 
1, & \text{if } \text{perfectly match}, \\
-0.5, & \text{if  } i_t \notin T_{\text{pred}}, \\
-1, & \text{otherwise}.
\end{cases}
\label{eq:reward}
\end{equation}

\paragraph{Critic}\label{sec:critic}
To enhance training stability, we introduce the critic to estimate the state-value function $V(s_t)$.
The input of $V(s_t)$ is a linearized state, and its output is a scalar representing the return (i.e., cumulative reward) when starting from state $s_t$. In the simplest case, the return is the chained sum of the rewards. Accordingly, one-step temporal difference (TD)~\citep{sutton1988learning} is often used to estimate $V(s_t)$, which is updated by the TD-target:
\begin{equation}
    G_t = r_t + \gamma V(s_{t+1}),
    \label{eq:sequence_Gt}
\end{equation}
where $\gamma$ is the discount factor. However, in structured reasoning, reasoning steps typically adhere to inherent tree~\citep{dalvi-etal-2021-explaining} or graph~\citep{DBLP:conf/iclr/Ribeiro0MZDKBRH23} structures, with the chained structure being merely a special case. Thus, Equation~\ref{eq:sequence_Gt} just describes the chained multi-step reasoning, which may not effectively capture the intricate logical dependencies between steps in structured reasoning.

Therefore, we propose the structure-based return, where the TD-target is expressed in a more general formulation: 
\begin{equation}
    \hat{G}_t = r_t + \gamma \frac{1}{|\mathcal{P}(s_t)|} \sum_{s_j\in \mathcal{P}(s_t)} V(s_{j}),
    \label{eq:structured_return}
\end{equation}
where $\mathcal{P}(s_t)$ represents the parent node of state $s_t$ in the entailment tree $T_{\text{pred}}$ or reasoning graph.
When $s_t \notin T_{\text{pred}}$, $\mathcal{P}(s_t) = s_{t+1}$. It can be seen that our structure-based return (Equation~\ref{eq:structured_return}) adapts to structured reasoning involving chained, tree-based and graph-based structured scenarios. Especially, entailment tree is a special case of the reasoning graph, in which each state typically has only one parent node, and thus Equation~\ref{eq:structured_return} degenerates into $\hat{G}_t = r_t + \gamma V(\mathcal{P}(s_t))$.
Furthermore, as shown in Figure~\ref{fig:equivalent_trajectory} (Appendix~\ref{sec:appendix_case_study}), for equivalent trajectories $s_1 \rightarrow s_2 \rightarrow s_3 \rightarrow  s_4$ and $s_2 \rightarrow s_1 \rightarrow s_3 \rightarrow  s_4$, previous method~\citep{liu-etal-2022-rlet} would assign different returns for state $s_1$ and $s_2$, even though they represent the same tree in the end.
Conversely, our method, by precisely delineating the intricate interdependencies between reasoning steps, consistently allocates the same return to any equivalent trajectories, thereby enhancing both stability and effectiveness.
\subsection{Optimization}
Our objective is to enhance the structured reasoning capabilities of the policy through RL.
To alleviate issues of training instability and sample inefficiency in RL~\citep{zhou-etal-2023-facilitating,roit-etal-2023-factually}, we employ the proximal policy optimization (PPO) algorithm~\citep{schulman2017proximal} to train the policy $\pi$ (parameterized by $\theta$), as follows:
\begin{equation}
    \begin{gathered}
    \mathcal{L}_{\pi} = \mathbb{E}_t[\min\Big( 
        \frac{\pi'_{\theta}(a_t|s_t)}{\pi'_{\theta_{old}}(a_t|s_t)}\hat{A}_t, \text{clip}\big(\frac{\pi'_{\theta}(a_t|s_t)}{\pi'_{\theta_{old}}(a_t|s_t)}, \\ 1-\epsilon, 1+\epsilon \big) \Big) \hat{A}_t + \beta \mathcal{E}(\pi'_{\theta})],
    \end{gathered}
    \label{eq:policy_loss}
\end{equation}
where $\pi'$ represents the probabilities normalized by Equation~\ref{eq:normalized_prob}, $\theta$ and $\theta_{old}$ are parameters of the new and old policies, $\epsilon$ is a hyperparameter defining the clipping range, $\beta$ is the entropy exploration coefficient, and $\mathcal{E}$ is the entropy bonus, which encourages sufficient exploration:
\begin{equation}
    \mathcal{E}(\pi'_{\theta}) = \mathbb{E}_{a_t\sim \pi_{\theta}}[-\log \pi'_{\theta}(a_t|s_t)].
    \label{eq:entropy}
\end{equation}
Futhermore, $\hat{A}_t$ is the estimate of the advantage function for state $s_t$, defined as follows:
\begin{equation}
    \hat{A}_t = \hat{G}_t - V(s_t).
    \label{eq:advantage}
\end{equation}

To accurately evaluate return and guide the policy towards better updates, we train the critic by minimizing the difference between its prediction and the TD-target:
\begin{equation}
    \mathcal{L}_V = \mathbb{E}_{t} \left[ (V(s_t) - \hat{G}_t)^2 \right].
    \label{eq:critic_loss}
\end{equation}

\paragraph{Supervised Warm-up}
Incorporating the supervised warm-up strategy before RL offers a relatively stable initial policy, which facilitates faster adaptation to the environment, particularly for complex reasoning tasks~\citep{DBLP:conf/iclr/RamamurthyABHSB23,wu-etal-2023-lilgym}.
Therefore, we convert the structured reasoning into single-step supervised data to warm up the policy as follows:
\begin{equation}
    \mathcal{L}_{\text{warmup}} = -\sum_{i}\log p(y_i|s_t,y_{<i}).
    \label{eq:supervised_warm_up}
\end{equation}
where $y$ is the golden action at $s_t$.

\section{Experiments}
\subsection{Datasets}
\paragraph{Tree-structured reasoning}
We conduct experiments on EntailmentBank~\citep{dalvi-etal-2021-explaining}, the first dataset that supports structured explanation with entailment trees.
Following~\citep{hong-etal-2023-faithful}, we also conduct experiments on EntailmentBankQA~\citep{tafjord-etal-2022-entailer}, whose objective is to reach the answer based on the entailment tree.

\paragraph{Graph-structured reasoning}
We conduct experiments on the STREET benchmark~\citep{DBLP:conf/iclr/Ribeiro0MZDKBRH23} to assess the performance of graph-structured reasoning.
Please refer to Appendix~\ref{sec:appendix_datasets_details} for more details about the dataset statistics.

\subsection{Baselines}
For EntailmentBank, we compare with single-pass methods, such as EntailmentWriter~\citep{dalvi-etal-2021-explaining}, and step-by-step methods including METGEN~\citep{hong-etal-2022-metgen}, IRGR~\citep{neves-ribeiro-etal-2022-entailment}, RLET~\citep{liu-etal-2022-rlet}, NLProofs~\citep{yang-etal-2022-generating} and FAME~\citep{hong-etal-2023-faithful}.
For EntailmentBankQA, we compare with Selection-Inference (SI)~\citep{DBLP:journals/corr/abs-2208-14271} and FAME~\citep{hong-etal-2023-faithful}.
For the STREET benchmark, we compare with the method proposed in~\citep{DBLP:conf/iclr/Ribeiro0MZDKBRH23}.
Furthermore, we conduct comparisons with GPT-4~\citep{openai2023gpt4} equipped with Chain-of-Thought (CoT)~\citep{NEURIPS2022_9d560961}, Tree of Thought (ToT)~\citep{yao2023tree} and ReAct~\citep{DBLP:conf/iclr/YaoZYDSN023}.




\subsection{Implementation Details}
For a fair comparison\footnote{Previous studies have consistently utilized T5-large as the base model. Despite the existence of more advanced generative models~\citep{du-etal-2022-glm,touvron2023llama}, using T5-large enables us to maintain a fair comparison.}, the policy is built with a T5-large model~\citep{10.5555/3455716.3455856}, while the critic is the encoder of T5-large combined with a MLP ($\tanh$ as the activation function).
For a supervised warm-up, we set a learning rate of 1e-5, a batch size of 16, and train the model for 20 epochs. For RL training, we set learning rate 2e-6 for both policy and critic, discounter factor $\gamma$ as 0.95, batch size as 3, buffer size as 12, buffer training epochs $N_K$ as 2, $\epsilon$ as 0.2, and $\beta$ as 1e-4.
More implementation details can be found in Appendix~\ref{sec:appendix_inplementation_details}.
\subsection{Evaluation Metrics}
For EntailmentBank, we evaluate $T_{\text{pred}}$ with the following dimensions: Leaves, Steps, Intermediates, and Overall AllCorrect.
For STREET benchmark, we evaluate the reasoning graphs with two dimensions: Answer Accuracy and Reasoning Graph Accuracy.
Note that Overall AllCorrect and Reasoning Graph Accuracy are extremely \textbf{strict} metrics, where any deviations will result in a score of 0.
More metrics details can be found in Appendix~\ref{sec:appendix_metric}.

\begin{table*}[ht]
\small
\centering
    \resizebox{0.99\linewidth}{!}{
\begin{tabular}{@{}clccccccc@{}}
\toprule
\multirow{2}{*}{\textbf{Task}} & \multicolumn{1}{c}{\multirow{2}{*}{\textbf{Method}}} & \multicolumn{2}{c}{\textbf{Leaves}} & \multicolumn{2}{c}{\textbf{Steps}}  & \multicolumn{2}{c}{\textbf{Intermediates}} & \textbf{Overall} \\
                               & \multicolumn{1}{c}{}                                 & F1               & AllCorrect       & F1               & AllCorrect       & F1                   & AllCorrect          & AllCorrect       \\ \midrule
\multirow{6}{*}{Task1}         & EntailmentWriter                                     & 98.7             & 84.1             & 50.0             & 38.5             & 67.6                 & 35.9                & 34.4             \\
                               & METGEN                                               & \textbf{100.0}   & \textbf{100.0}   & \underline{57.9} & 42.1             & \underline{71.3}     & 39.2                & 37.0             \\
                               & IRGR                                                 & 97.6             & 89.4             & 50.2             & 36.8             & 62.1                 & 31.8                & 32.4             \\
                               & RLET                                                 & \textbf{100.0}   & \textbf{100.0}   & 54.6             & 40.7             & 66.9                 & 36.3                & 34.8             \\
                               & NLProofS                                             & 97.8             & 90.1             & 55.6             & \underline{42.3} & \textbf{72.4}        & \underline{40.6}    & \underline{38.9} \\ \cmidrule(l){2-9} 
                               & \modelname (Ours)                                    & \textbf{100.0}   & \textbf{100.0}   & \textbf{67.6}    & \textbf{52.6}    & 70.3                 & \textbf{42.6}       & \textbf{40.6}    \\ \midrule
\multirow{6}{*}{Task2}         & EntailmentWriter                                     & 83.2             & 35.0             & 39.5             & 24.7             & 62.2                 & 28.2                & 23.2             \\
                               & METGEN                                               & 83.7             & 48.6             & 41.7             & 30.4             & 62.7                 & 32.7                & 28.0             \\
                               & IRGR                                                 & 69.9             & 23.8             & 30.5             & 22.3             & 47.7                 & 26.5                & 21.8             \\
                               & RLET                                                 & 81.0             & 39.0             & 38.5             & 28.4             & 56.3                 & 28.6                & 25.7             \\
                               & NLProofS                                             & \textbf{90.3}    & \textbf{58.8}    & \underline{47.2} & \underline{34.4} & \textbf{70.2}        & \underline{37.8}    & \underline{33.3} \\ \cmidrule(l){2-9} 
                               & \modelname (Ours)                                    & \underline{86.4} & \underline{53.5} & \textbf{56.8}    & \textbf{39.7}    & \underline{66.3}     & \textbf{38.3}       & \textbf{34.7}    \\ \midrule
\multirow{10}{*}{Task3}        & EntailmentWriter                                     & 35.7             & 2.9              & 6.1              & 2.4              & 33.4                 & 7.7                 & 2.4              \\
                               & METGEN                                               & 34.8             & 8.7              & 9.8              & 8.6              & 36.7                 & 20.4                & 8.6              \\
                               & IRGR                                                 & 45.6             & 11.8             & 16.1             & 11.4             & 38.8                 & \underline{20.9}    & 11.5             \\
                               & RLET                                                 & 38.3             & 9.1              & 11.5             & 7.1              & 34.2                 & 12.1                & 6.9              \\
                               & NLProofS                                             & 43.2             & 8.2              & 11.2             & 6.9              & 42.9                 & 17.3                & 6.9              \\
                               & FAME                                                 & 43.4             & \textbf{13.8}    & \underline{16.6} & \underline{12.4} & 40.6                 & 19.9                & \underline{11.9} \\
                               & GPT4-CoT                                             & 44.1             & 12.1             & 15.4             & 10.8             & 43.1                 & 20.6                & 10.8             \\
                               & GPT4-ToT                                             & 43.3             & 12.0             & 15.8             & 11.0             & 43.9                 & 20.0                & 11.0             \\
                               & GPT4-ReAct                                           & \underline{45.8} & 12.9             & 14.1             & 10.5             & \underline{43.5}     & \textbf{21.5}       & 10.5             \\ \cmidrule(l){2-9} 
                               & \modelname (Ours)                                    & \textbf{47.1}    & \textbf{13.8}    & \textbf{17.4}    & \textbf{12.9}    & \textbf{45.1}        & 18.8                & \textbf{12.9}    \\ \bottomrule
\end{tabular}
}
\caption{Experiment results on EntailmentBank. Bold and underlined texts highlight the best method and the runner-up. RLET is based on DeBERTa-large~\citep{DBLP:conf/iclr/HeGC23}, while all other methods are based on T5-large. All baseline results come from published papers. We use the \texttt{gpt-4-1106-preview} version for GPT-4.}
\label{tab:entailmentbank}
\end{table*}

\section{Result Analysis}
\subsection{Structured Reasoning}
\paragraph{EntailmentBank}
As shown in Table~\ref{tab:entailmentbank}, our \modelname outperforms all baseline methods on the most strict metric, "Overall AllCorrect", across all three tasks.
Specifically, our method achieves an absolute improvement of 1.7\%/1.4\%/1.0\% in Task 1/2/3 compared to the strongest baseline.
The steps dimension, i.e., premises selection, is the core of EntailmentBank\footnote{A comprehensive error analysis is detailed in Appendix~\ref{sec:appendix_error_analysis}.}, contributing to enhancing the accuracy of both leaves and intermediates dimensions, thereby improving the overall AllCorrect metric.
(1) Compared to SOTA supervised methods, such as NLProofs and FAME, our method exhibits significant advantages in the steps dimension.
This demonstrates that focusing solely on isolated single-step reasoning through supervised learning may yield suboptimal solutions in intricate structured reasoning tasks, even though employing advanced planning algorithms, such as Monte-Carlo planning in FAME.
(2) Compared to the SOTA RL-based method, our method outperforms RLET by 5.8\%/9.0\%/6.0\% in Task 1/2/3.
Our method employs a generative model as the policy to circumvent the issue of enumerating actions, facilitating the policy's understanding of structured reasoning tasks (generating potential actions by itself).
Moreover, our proposed structure-based return more effectively captures the tree-structured logical dependencies between steps and can assign stable returns for equivalent trajectories, which significantly improves reasoning abilities.
Subsequent ablation studies will further demonstrate this.
(3) Compared to GPT-4 with CoT, ToT, and ReAct, our method achieves an absolute improvement of 1.9\% in Task 3.
Although GPT-4 exhibits outstanding reasoning capabilities surpassing many other baselines, its performance relies on a vast number of parameters.
Details about the prompts of GPT-4 can be found in Appendix~\ref{sec:appendix_GPT4}.

\begin{table}[t]
\centering
    \resizebox{0.7\linewidth}{!}{
\begin{tabular}{lcc}
\hline
Method           & Task 1           & Task 2           \\ \hline
SI+Halter        & 72.4             & 55.9             \\
SI+Halter+Search & 83.2             & 72.9             \\
FAME             & \underline{91.5} & \underline{78.2} \\ \hline
\modelname (Ours)             & \textbf{92.7}    & \textbf{85.6}    \\ \hline
\end{tabular}
}
\caption{Experiment results on the EntailmentBankQA. SI is based on Chinchilla-7B~\citep{DBLP:journals/corr/abs-2203-15556}.}
\label{tab:entailmentbankQA}
\end{table}

\paragraph{EntailmentBankQA}
Following~\citet{DBLP:journals/corr/abs-2208-14271}, we introduce the halter module to generate answers based on $T_{\text{pred}}$ and substitute hypothesis with question and option during the reasoning process.
As illustrated in Table~\ref{tab:entailmentbankQA}, our method surpasses FAME by an absolute margin of 1.2\%/7.4\% in Task 1/2.
While both FAME and SI are supervised methods, FAME significantly outperforms SI by enhancing the model's reasoning and exploration capabilities through Monte-Carlo planning.
However, our method enhances the structured reasoning capabilities of the policy rather than focusing solely on single-step reasoning, which can significantly improve the quality of the entailment tree to aid in answering, especially in complex reasoning environments.
\begin{table}[t]
\centering
    \resizebox{\linewidth}{!}{
\begin{tabular}{@{}lcccc@{}}
\toprule
\textbf{Method} & \textbf{SCONE}   & \textbf{GSM8K}   & \textbf{AQUA-RAT} & \textbf{AR-LSAT} \\ \midrule
\multicolumn{5}{c}{\textbf{Answer Accuracy}}                                                 \\ \midrule
STREET          & \underline{69.6} & 10.4             & 28.7              & 28.0             \\
GPT4~\dag            & 66.0             & \textbf{94.0}    & \textbf{78.0}     & \underline{32.0} \\
SEER (Ours)     & \textbf{72.4}    & \underline{21.4} & \underline{37.6}  & \textbf{33.5}    \\ \midrule
\multicolumn{5}{c}{\textbf{Reasoning Graph Accuracy}}                                        \\ \midrule
STREET          & \underline{60.0} & 0.7              & 0.0               & 0.0              \\
GPT4~\dag            & 32.0             & \underline{10.0} & \underline{4.0}  & \underline{2.0}  \\
SEER (Ours)     & \textbf{64.8}    & \textbf{13.4}    & \textbf{8.1}     & \textbf{7.2}    \\ \bottomrule
\end{tabular}
}
\caption{Experiment results on STREET benchmark. \dag~indicates we recorded the best results in CoT, ToT, and ReAct for brevity.}
\label{tab:STREET}
\end{table}
\paragraph{STREET}
As shown in Table~\ref{tab:STREET}, compared to GPT-4, our method has achieved absolute improvements of 4.8\%/3.4\%/4.1\%/5.2\% across various datasets, although the Reasoning Graph Accuracy is a very strict metric~\citep{DBLP:conf/iclr/Ribeiro0MZDKBRH23}.
While GPT-4 excels at answering questions (far surpassing other methods), its parameter is thousands of times greater than other methods.
Moreover, during the reasoning process, GPT-4 is prone to hallucinations~\citep{DBLP:journals/corr/abs-2309-05922}, resulting in poor performance in structured reasoning, particularly evident in the "Reasoning Graph Accuracy" metric.
Since SCONE contains sufficient data as well as similar QA and reasoning patterns, we observe that the STREET method would outperform GPT-4 on SCONE.
However, by obtaining high-quality reasoning graphs, our method achieves absolute improvements of 2.8\%/11.0\%/8.9\%/5.5\% compared to the STREET method, significantly improving answer accuracy and trustworthiness.
In reasoning graphs, a state may have multiple parent nodes.
Our structure-based return (Equation~\ref{eq:structured_return}) still precisely describes the cumulative reward for each state, thereby facilitating reasoning performance in graph-structured reasoning.

\subsection{Cross-dataset Performance}
\begin{table}[t]
\centering
    \resizebox{\linewidth}{!}{
\begin{tabular}{lcccc}
\hline
\multirow{2}{*}{Method} & \multicolumn{2}{c}{eQASC}             & \multicolumn{2}{c}{eOBQA}             \\
                        & P@1               & NDCG              & P@1               & NDCG              \\ \hline
EntailmentWriter        & 52.48             & 73.14             & 69.07             & 89.05             \\
EntailmentWriter-Iter   & 52.56             & 73.28             & 72.15             & 90.19             \\
METGEN                  & \underline{55.81} & 74.19             & 74.89             & 90.50             \\
FAME                    & 53.36             & 79.64             & 73.09             & 89.32             \\
GPT-4                   & 54.00             & \underline{88.82} & \textbf{85.36}    & \underline{91.19} \\ \hline
\modelname (Ours)       & \textbf{60.33}    & \textbf{89.76}    & \underline{77.50} & \textbf{94.62}    \\ \hline
\end{tabular}
}
\caption{Cross-dataset performance on the eQASC and eOBQA.}
\label{tab:cross_dataset}
\end{table}
To evaluate the generalization performance, we conduct cross-dataset experiments on eQASC and eOBQA\footnote{More details about the setting of eQASC and eOBQA can be found in Appendix~\ref{sec:appendix_datasets_details}.}~\citep{jhamtani-clark-2020-learning}.
We apply the policy of Task 2 for selection without training on eQASC or eOBQA.
As illustrated in Table~\ref{tab:cross_dataset}, our method exhibits significant superiority in cross-dataset generalization.
Compared to supervised methods, our \modelname, following the inherent structural nature of entailment trees, can better capture the logical dependencies between reasoning steps, which can effectively promote the generalization ability of the policy.
The experimental results further validate the effectiveness of our method.

\subsection{Ablation Studies}
\begin{table}[t]
\centering
    \resizebox{\linewidth}{!}{
\begin{tabular}{@{}lcccc@{}}
\toprule
Method             & Leaves        & Steps         & Intermediates & Overall       \\ \midrule
\modelname (Ours)               & \textbf{13.8} & \textbf{12.9} & \textbf{18.8} & \textbf{12.9} \\
w/o redundant      & 13.2          & 12.6          & 18.5          & 12.3          \\
w/o structure-based return & 12.9          & 11.7          & 18.5          & 11.1          \\
w/o RL             & 10.2          & 9.4          & 17.1          & 9.1          \\ \bottomrule
\end{tabular}
}
\caption{Ablation study of each component.}
\label{tab:ablation}
\end{table}

\begin{figure}[t]
    \begin{minipage}[c]{1\linewidth}
        \centering

        \begin{subfigure}[b]{0.49\linewidth}
            \centering
            \includegraphics[width=\linewidth]{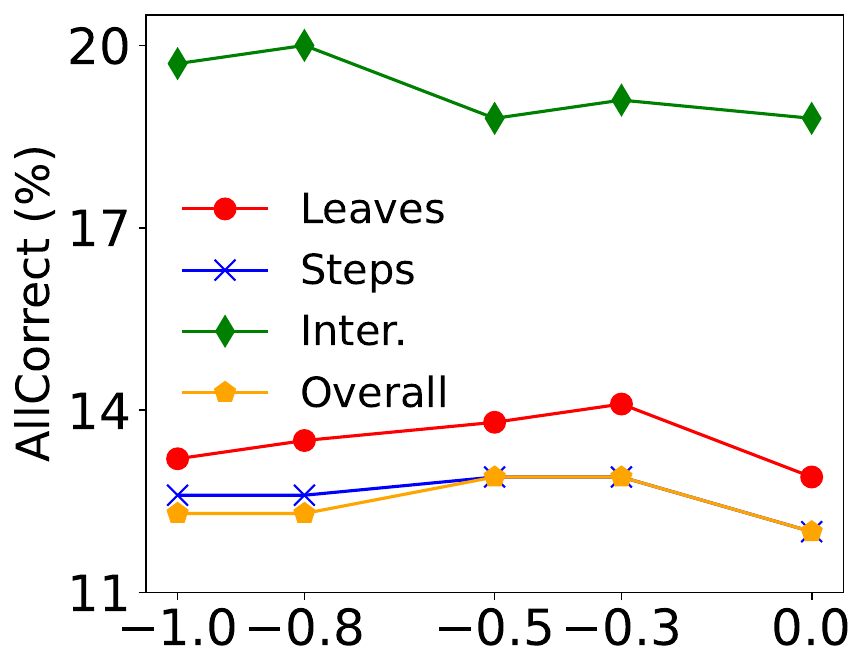} 
            \caption{$r_{\text{redundant}}$}
        \end{subfigure}
        \begin{subfigure}[b]{0.49\linewidth}
            \centering
            \includegraphics[width=\linewidth]{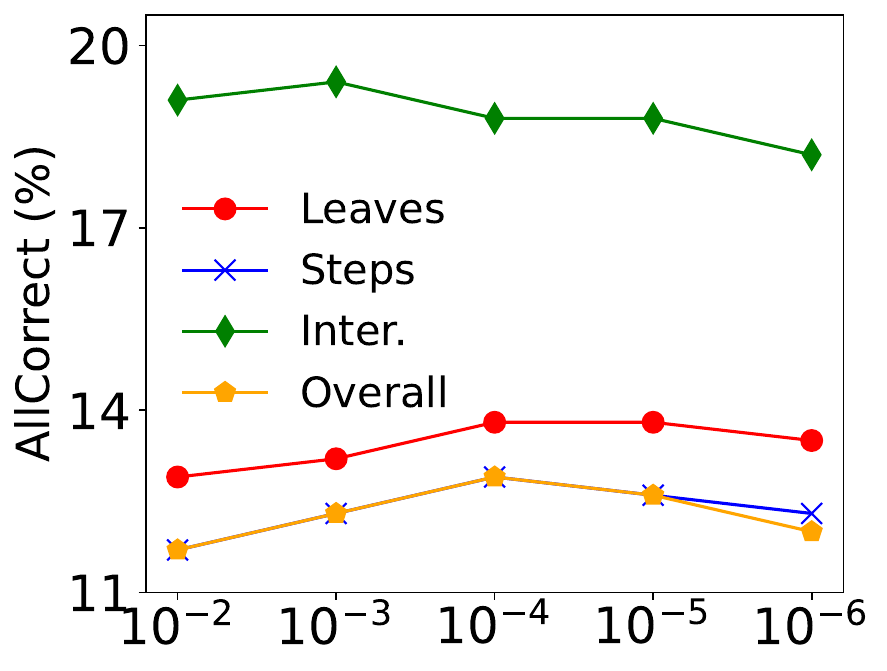} 
            \caption{$\beta$}
        \end{subfigure}

        \caption{Parameter sensitivity analysis.}
        \label{fig:param_sensitive}
    \end{minipage}%
\end{figure}

To evaluate the contribution of each component, we conduct extensive ablation studies.
As shown in Table~\ref{tab:ablation}, we investigate three different variations of \modelname in Task 3 of EntailmentBank: (1) \textbf{w/o redundant} neglects redundant steps by assigning a reward of -1.
(2) \textbf{w/o structure-based return} removes the structure-based return and calculates it using the chained sum of rewards (Equation~\ref{eq:sequence_Gt}).
(3) \textbf{w/o RL} removes the RL phase, relying solely on supervised warm-up.
We discover that overlooking redundant steps may potentially inhibit the exploration of policy, leading to a performance decline.
In addition, the results shown in Table~\ref{tab:ablation} also demonstrate that removing the structure-based return severely affects the performance.
It not only adequately addresses the equivalent trajectory problems, but also elegantly captures the logical relationships inherent in entailment trees, which is crucial for structured reasoning.
Furthermore, it can be seen that removing the RL phase reduces performance by 3.8\% of Overall Allcorrect, which is a significant impact for this strict metric.
This indicates that relying solely on supervised learning may overlook the logical relationships in structured reasoning, thereby falling into suboptimal solutions.

\subsection{Parameter Sensitivity Analysis}
\label{sec:param_sensitive}
As illustrated in Figure~\ref{fig:param_sensitive}, we further investigate the impact of $r_{\text{redundant}}$ and $\beta$ on the performance in Task 3.
We observe that compared to treating redundant and erroneous steps equally ($r_{\text{redundant}}=-1$), not penalizing ($r_{\text{redundant}}=0$) may have more detrimental effects, which allows for unrestricted exploration.
Moreover, a suitable $\beta$ (the coefficient of entropy bonus) is crucial for performance enhancement, as it encourages the policy to break away from the "stereotypes" of supervised warm-up.

\section{Conclusions}
We propose \modelname, a novel approach that facilitates structured reasoning and explanation via RL.
To our knowledge, \modelname is the first general framework capable of enhancing chained, tree-based, and graph-based structured reasoning.
Our structure-based return precisely delineates the hierarchical and branching structure inherent in structured reasoning, effectively facilitating reasoning ability.
Furthermore, \modelname employs a generative model to represent the policy and refines the reward function, ingeniously circumventing the limitations of existing works.
Comprehensive experimental results demonstrate that \modelname significantly outperforms state-of-the-art methods and exhibits outstanding cross-dataset generalization performance.

\section*{Limitations}
Although our method has achieved excellent performance in structured reasoning and explanation, there remains one issue that deserves further exploration for future work: how to perform structured reasoning in the context of multimodal data.
This includes combining content from images, tables, or audio data, a form of multimodal structured reasoning that is increasingly prevalent and demanding in real-world scenarios.
In future work, we plan to extend our \modelname to accommodate multimodal scenarios.

\section*{Ethics Statement}
This work focuses primarily on structured reasoning and explanation problems, and its contributions are entirely methodological. Therefore, this work does not have direct negative social impacts.
For the experiments, we have open-sourced the code and utilized openly available datasets commonly used in previous research, without any sensitive information to our knowledge.
The authors of this work adhere to the ACL ethical guidelines, and the application of this work does not present any apparent issues that may lead to ethical risks.


\section*{Acknowledgements}
This work is supported by the National Key R\&D Program of China (NO.2022ZD0160102). Chao Yang is supported by the Shanghai Post-doctoral Excellent Program (Grant No. 2022234).

\bibliography{custom}
\bibliographystyle{acl_natbib}

\newpage
\clearpage

\appendix
\section{Algorithm Details}
\label{sec:appendix_algorithm_details}
\begin{algorithm}[ht]
\caption{The training process of \modelname}
\label{alg:train}
\LinesNumbered 
\KwIn{Structured reasoning dataset $\mathcal{D}$; Training epochs $N_{\text{warmup}}$, $N$ and $N_K$; batch size $b_{\text{warmup}}$ and $b_{\text{mini}}$; 
}
\KwOut{The optimal parameter of policy}
\tcc{(1) Supervised Warm-up phase}
Initialise policy parameters $\pi_{\theta}$ \\
Convert $\mathcal{D}$ into single-step data $\mathcal{D}_{step}$ \\
\For{epoch $=1$ to $N_{\text{warmup}}$}{
    \For{i $=1$ to  $|\mathcal{D}_{step}|/b_{\text{warmup}}$}{
    sample minibatch from $\mathcal{D}_{\text{step}}$ \\
    update parameters $\pi_{\theta}$ by Eq.~\ref{eq:supervised_warm_up}
    }
}
\tcc{(2) RL phase}
Initialize the critic parameters $V$\\
\For{epoch $=1$ to $N$}{
    Initialise training buffer $\mathcal{B} \leftarrow \varnothing$\\
    \tcp{Filling the replay buffer}
    \While{$\mathcal{B}$ not full}{
        sample $\{h,X, T_{\text{gold}}\}$ from $\mathcal{D}$\\
        collect trajectory $\tau$ via $\pi_{\theta}$ \\
        assign a reward for each step in $\tau$ \\
        fill buffer $\mathcal{B}$ with $\{s_t, a_t, r_t\}$ from $\tau$ \\
    }
    \tcp{Performing k-epoch updates per buffer}
    \For{$\text{epoch}_{k}$ $=1$ to $N_K$}{
        sample $\{(s_t, a_t, r_t)\}_{b_{\text{mini}}}$ from $\mathcal{B}$\\
        compute $\mathcal{E}(\pi'_{\theta})$ and $\hat{A}_t$ by Eqs.~\ref{eq:entropy}, \ref{eq:advantage} \\ 
        update policy $\pi_{\theta}$ by Eq.~\ref{eq:policy_loss}\\
        update critic $V$ by Eq.~\ref{eq:critic_loss}\\
    }
}
\end{algorithm}
Algorithm~\ref{alg:train} describes the overall training process of our proposed \modelname in detail, which primarily consists of two phases: supervised warm-up and reinforcement learning (RL).
In the supervised warm-up phase, the structured reasoning is first decomposed into single-step reasoning data (Line 2).
Then, we employs supervised learning to guide the policy $\pi_{\theta}$ to quickly adapt to the complex reasoning environments (Lines 3-6).
This is particularly beneficial when the number of parameters in the policy is relatively small~\citep{akyurek-etal-2023-rl4f,liu-etal-2023-one}.
In the RL phase, we initially populate the replay buffer $\mathcal{B}$ through the policy $\pi_{\theta}$ (Lines 9-13).
Then, we update the parameters of the policy and the critic using the buffer data. To improve sample efficiency, $N_K$ updates are performed for each replay buffer (Lines 14-18).

For the inference process, we only need to use the policy (without the critic) for structured reasoning.
Specifically, as illustrated in the trajectory rollout of Figure~\ref{fig:framework}, we update the state by the policy and the entailment module.
Then, we end the reasoning process until the stopping criteria are satisfied.
Finally, we backtrack to construct the entire structured explanation, taking the last intermediate conclusion as the hypothesis for entailment tree (or the answer for the STREET benchmark).

\section{Datasets Details}
\label{sec:appendix_datasets_details}
\begin{table*}[t]
\centering
    \resizebox{0.99\linewidth}{!}{
\begin{tabular}{lccccc}
\hline
\textbf{Task Name} & \textbf{Task Domain} & \textbf{\# Questions} & \textbf{\# Reasoning Steps} & \textbf{Reasoning Type} & \textbf{Answer Type} \\ \hline
EntailmentBank     & Science              & 1,840                 & 5,881                       & Tree-structured         & /                    \\
EntailmentBankQA (ARC)   & Science              & 1,840                 & 5,881                       & Tree-structured         & 4-Way MC             \\
SCONE              & Processes            & 14,574                & 130,482                     & Graph-structured        & State Pred.          \\
GSM8K              & Math                 & 1,030                 & 4,666                       & Graph-structured        & Number               \\
AQUA-RAT           & Math                 & 1,152                 & 7,179                       & Graph-structured        & 5-Way MC             \\
AR-LSAT            & Logic                & 500                   & 2,885                       & Graph-structured        & 5-Way MC             \\ \hline
\end{tabular}
}
\caption{
Datasets Statistics of Structured Reasoning.
}
\label{tab:datasets_statistics}
\end{table*}
\paragraph{Datasets of Structured Reasoning}
Table~\ref{tab:datasets_statistics} describes the statistics of datasets in detail.
In the answer types, “K-Way MC” stands for multiple choice answer with K options.

EntailmentBank~\citep{dalvi-etal-2021-explaining} comprises 1,840 expert-annotated entailment trees with an average of 7.6 nodes spanning across 3.2 entailment steps.
The facts are derived from the WorldTree V2 corpus~\citep{xie-etal-2020-worldtree}.
Based on different facts $X$, there are three progressively more challenging tasks: \textbf{Task1 (no-distractor)}, \textbf{Task2 (distractor)} and \textbf{Task3 (full-corpus)}.
For GPT-4, we employ all the data in Task 3 from EntailmentBank to evaluate its performance.
EntailmentBank was originally designed for post-hoc tree reconstruction tasks instead of QA, \citet{tafjord-etal-2022-entailer} converted it into EntailmentBankQA where the task is to choose the correct answer given multiple choice options rather than deriving hypothesis $h$.

To construct the STREET benchmark, \citet{DBLP:conf/iclr/Ribeiro0MZDKBRH23} standardized many QA datasets, such as ARC~\citep{DBLP:journals/corr/abs-1803-05457}, SCONE~\citep{long-etal-2016-simpler}, GSM8K~\citep{cobbe2021training}, AQUA-RAT~\citep{ling-etal-2017-program} and AR-LSAT~\citep{zhong2021ar}, in the graph-structured explanation format, where the tasks are converted into answering the question based on the predicted reasoning graphs.
Please note that ARC in STREET is congruent with Task 1 of EntailmentBankQA~\citep{DBLP:conf/iclr/Ribeiro0MZDKBRH23}, hence, we do not repeat the experiment for this task in Table~\ref{tab:STREET}. 
Due to the high cost of GPT-4, we randomly sample 50 instances from each dataset in the STREET benchmark to evaluate GPT-4's performance.

\paragraph{Datasets of Cross-dataset Experiments}
To evaluate the generalization performance of our method, following~\citet{hong-etal-2022-metgen}, we conduct cross-dataset experiments on eQASC and eOBQA~\citep{jhamtani-clark-2020-learning},
which collect \textit{one-step} entailment trees for questions from QASC~\citep{Khot_Clark_Guerquin_Jansen_Sabharwal_2020} and OpenBookQA~\citep{mihaylov-etal-2018-suit}, respectively.
The goal of this task is to select valid one-step trees from the candidate set and evaluate the results with P@1 and NDCG metrics~\citep{hong-etal-2022-metgen}.
Questions with no valid tree are filtered.
The candidate sets for eQASC and eOBQA are composed of 10 and 3 sentences respectively.

\section{Implementation Details}
\label{sec:appendix_inplementation_details}
\subsection{Stopping criteria}
\label{sec:appendix_stop_condition}
For a fair comparison, we use the T5-large model to represent the policy.
However, we observe that T5-large tends to perform "\texttt{Reason}" actions more frequently, which is caused by the smaller number of model parameters and the issue of having only a few "\texttt{End}" instances.
Moreover, unlike GPT-4, T5-large is less able to recognize when a hypothesis has been inferred and when to stop.
Therefore, we attach two extra stopping criteria in addition to the "\texttt{End}" action: 
(1) The semantic similarity between the intermediate conclusion and the hypothesis exceeds a predefined threshold, i.e., BLEURT($i_*$, $h$) > 1.
(2) Exceeding the maximum number of reasoning steps (set to 20 in this paper).

\subsection{Alignment algorithm}
\label{sec:appendix_allignment}
Following~\citep{dalvi-etal-2021-explaining}, we evaluate the intermediate steps based on Jaccard similarity.
Specifically, the intermediate nodes $i_*$ in $T_{\text{pred}}$ are aligned with the intermediate nodes in $T_{\text{gold}}$ that have the maximum Jaccard similarity.
If the Jaccard similarity between the intermediate node in $T_{\text{pred}}$ and all intermediate nodes in $T_{\text{gold}}$ is zero, it is aligned with "NULL".
Note that only the intermediate node that is perfectly matched with a node in $T_{\text{gold}}$, i.e., the Jaccard similarity is 1, is considered as a correct step.
Figure~\ref{fig:reward_process} provides a detailed illustration of this process.
The alignment process is similar in the reasoning graphs~\citep{DBLP:conf/iclr/Ribeiro0MZDKBRH23}.

\subsection{Retriever for Task 3}
In Task 3 of EntailmentBank, first, it is necessary to retrieve relevant sentences from the corpus~\citep{dalvi-etal-2021-explaining}.
The research focus of this paper is to enhance the structured reasoning ability of the policy.
Therefore, we directly adopt the retriever and its associated parameters proposed in previous work~\citep{hong-etal-2023-faithful}, which is based on the all-mpnet-base-v2 model~\citep{reimers-2019-sentence-bert}.
For a fair comparison, we retrieve the top 25 most relevant sentences as $X$ for Task 3.

\subsection{Entailment Module}
The entailment module is also based on the T5-large model, taking premises as input and generating intermediate conclusions.
Our primary focus is to enhance the structured reasoning ability of the policy through RL, therefore, we directly employ the entailment module that has already been trained in previous work~\citep{hong-etal-2023-faithful}, which also aids in a fair comparison.

\subsection{Halter Module}
In EntailmentBankQA, we employ the Halter module~\citep{DBLP:journals/corr/abs-2208-14271} to answer questions based on the predicted entailment trees.
In this paper, the Halter module is built upon the T5-large model.
The module is trained with a learning rate of 1e-5 and a batch size of 16.

\subsection{Entailment Tree Construction}
\label{sec:appendix_tree_construction}
To evaluate the correctness of each reasoning step, we have to reconstruct the trajectory into an entailment tree $T_{\text{pred}}$ and compare it with $T_{\text{gold}}$.
Figure~\ref{fig:reward_process} illustrates this reconstruction process.
We consider the last intermediate conclusion as the hypothesis and then construct the predicted entailment tree based on the reasoning relationship of each step. 
The reconstruction process is similar in the reasoning graphs~\citep{DBLP:conf/iclr/Ribeiro0MZDKBRH23}.

\subsection{Running time}
\label{sec:appendix_runing_time}
In our experimental setting, the average training time per entailment tree in \modelname is 6.98 seconds, and the average inference time per entailment tree in \modelname is 3.91 seconds. As reported in their papers, the inference time per entailment tree in RLET~\citep{liu-etal-2022-rlet} and FAME~\citep{hong-etal-2023-faithful} are 9.34 seconds and 30.77 seconds, respectively. 
FAME leverages Monte-Carlo planning, necessitating the exploration of numerous nodes to enhance the reasoning capability of the policy, thus requiring considerable computational time. Our proposed \modelname significantly surpasses FAME in terms of both efficiency and effectiveness.

\subsection{Experiment Environments}
All experiments were conducted on Ubuntu 22.04 equipped with NVIDIA A100 GPUs.
Our code mainly depends on python 3.10\footnote{\url{https://www.python.org/}} and PyTorch 2.0.1\footnote{\url{https://pytorch.org/}}.
The pre-trained language models are derived from \texttt{HuggingFace Transformers}\footnote{\url{https://huggingface.co/}}.

\subsection{Details of Reasoning Graphs}
For the reasoning graphs in the STREET Benchmark, the implementation details are slightly different from the entailment trees.
In the reasoning graphs, reasoning steps may possess multiple parent nodes, and a fact ($x_*$) or intermediate conclusion ($i_*$) may be utilized multiple times~\citep{DBLP:conf/iclr/Ribeiro0MZDKBRH23}.
Therefore, in the reasoning graph, we refrain from incorporating previously used premises into $U_t$, instead continually expanding the candidate sentence set $C_t$ through newly derived intermediate conclusions.
In other words, the state in the reasoning graphs is updated according to the following rules:
$P_{t+1} = P_{t} \cup \{\text{\textless premises\textgreater} \rightarrow i_t\}$, $C_{t+1} = \{X \cup I_{t+1}\}$, and $I_{t+1} = I_t \cup \{i_t\}$.

\subsection{Other Implementation Details}
For GPT-4, we set the temperature to 0.7. For Tree of Thought, we set $b=5$ candidates at each step, and then vote to select the optimal action.
Details regarding the prompts of CoT, ToT, and ReAct can be found in Appendix~\ref{sec:appendix_GPT4}.
For all baselines, we obtain the optimal results based on experimental results or hyperparameter settings derived from the original papers.
For our method, we initialize the critic with the encoder of the warm-up policy to expedite the convergence of the critic and facilitate policy updates.
The hidden layer dimension of the MLP in the critic is set to 512.

\section{Metrics Details}
\label{sec:appendix_metric}
For EntailmentBank, we follow~\citep{dalvi-etal-2021-explaining} and evaluate the entailment tree $T_{\text{pred}}$ using three dimensions:

\noindent $\bullet$ \textbf{Leaves}: To evaluate the leaf nodes of $T_{\text{pred}}$, we compute F1 by comparing $X_{\text{pred}}$ with $X_{\text{gold}}$.

\noindent $\bullet$ \textbf{Steps}: To evaluate the structural correctness of each step, we compare all steps between $T_{\text{pred}}$ and $T_{\text{gold}}$ and then compute F1.
A predicted step is considered structurally correct if its premises (e.g., $x_*$, $i_*$) exactly match the gold premises.

\noindent $\bullet$ \textbf{Intermediates}:
To evaluate the intermediate conclusions, we compare the aligned intermediate conclusions and then compute F1.
A predicted intermediate conclusion is deemed correct if the BLEURT score~\citep{sellam-etal-2020-bleurt} exceeds 0.28.

For each dimension, the AllCorrect score is 1 if F1 is 1, otherwise 0.
Given the above scores, we employ the \textbf{Overall AllCorrect} metric to comprehensively evaluate $T_{\text{pred}}$, which takes a value of 1 if and only if all leaves, steps, and intermediates are correct.
Note that this is an extremely \textbf{strict} metric, where any deviations in $T_{\text{pred}}$ will result in a score of 0.

For the STREET benchmark, we follow~\citep{DBLP:conf/iclr/Ribeiro0MZDKBRH23} and adopt two metrics, namely, the answer to the question and the quality of the reasoning graphs, to evaluate different methods.

\noindent $\bullet$ \textbf{Answer Accuracy:} 
This metric measures the ability to correctly answer questions.
The answer accuracy serves as an upper bound for other metrics, as any reasoning graph generated with incorrect answers is also labeled as incorrect.

\noindent $\bullet$ \textbf{Reasoning Graph Accuracy:} 
This metric compares the predicted reasoning graph and the golden reasoning graph from the aspects of the graph structure and the content of intermediate conclusions.
Please note that this is a stringent metric, with minor deviations from the golden reasoning graph resulting in the prediction being incorrect.

\section{Illustrations and Case Study of \modelname}
\label{sec:appendix_case_study}
\begin{figure*}[t]
    \centering
    \includegraphics[width=\linewidth]{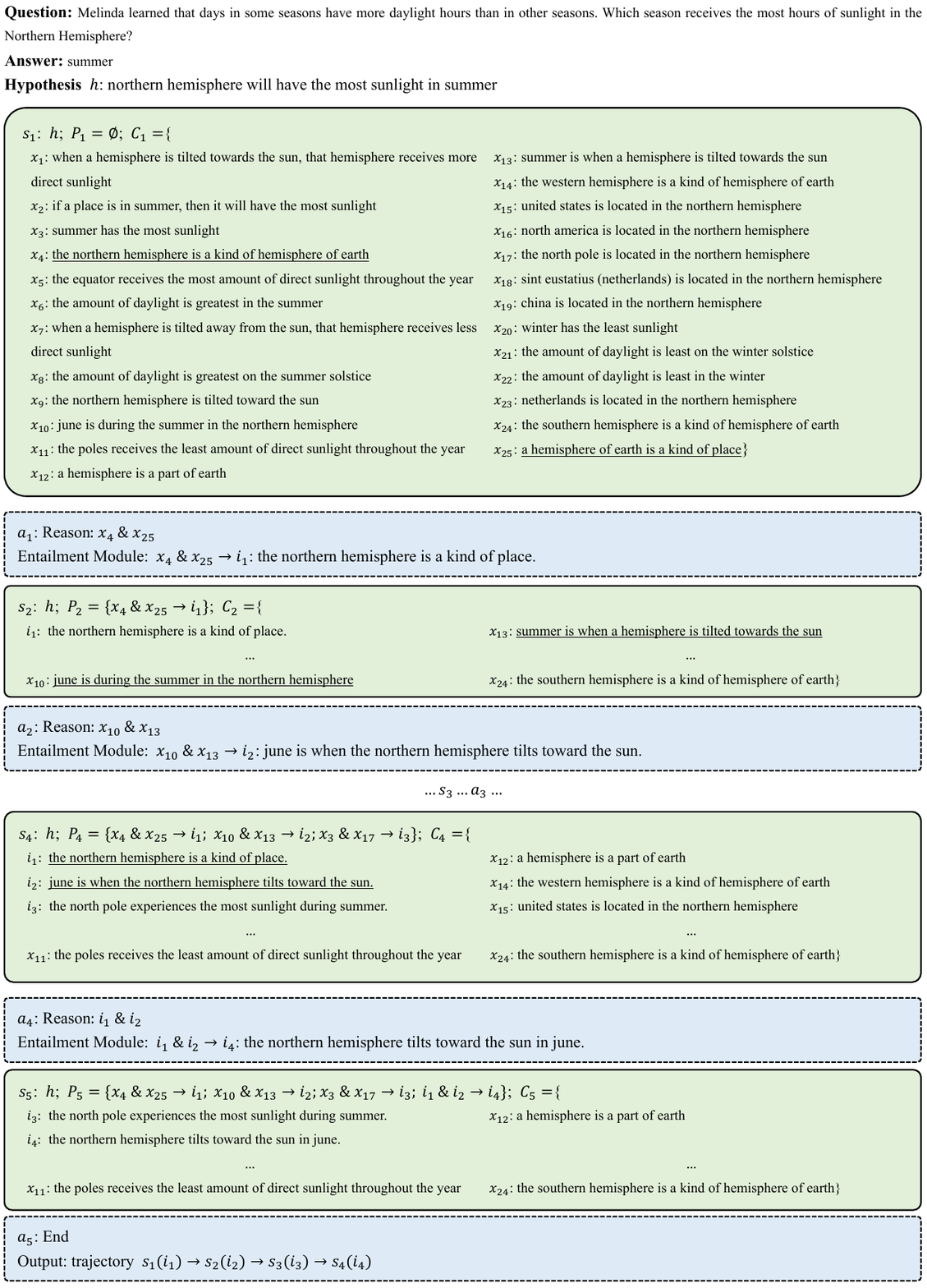}
    \caption{An illustration of the reasoning process of \modelname. Note that $a_1$ is a correct step, $a_2$ and $a_4$ are erroneous steps, and $a_3$ is a redundant step. We start from the initial state $s_1$ where existing entailment steps $P_1=\varnothing$ and candidate sentences $C_1=X$. In each step, we sample an action and update the state until the reasoning is done. For the "\texttt{Reason}" action, we sent the premises to the entailment module. The new conclusion is added to the $C$, the premises is removed from $C$ and the entailment step is added to the $P$. For the "\texttt{End}" action, we end the reasoning process and output the trajectory.}
    \label{fig:case_study}
\end{figure*}

\begin{figure*}[t]
    \centering
    \includegraphics[width=1\linewidth]{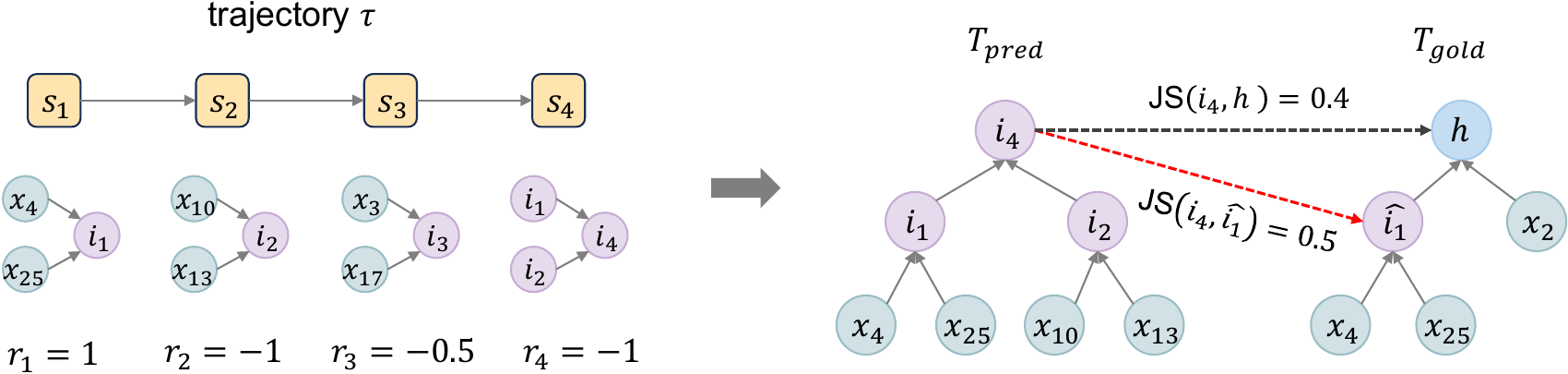}
    \caption{An illustration of the reward and alignment process of \modelname. Each reasoning step is a subtree (similarly, each reasoning step is a subgraph in the reasoning graph~\citep{DBLP:conf/iclr/Ribeiro0MZDKBRH23}). (1) We construct $T_{\text{pred}}$ using the last intermediate conclusion ($i_4$ in this example) as the hypothesis. (2) We calculate the Jaccard similarity between the intermediate node ($i_*$) in $T_{\text{pred}}$ and each golden intermediate node in $T_{\text{gold}}$ ($\hat{i}_1$ and $h$ in this example), and align with the maximum Jaccard similarity. In this example, $i_1$ is aligned with $\hat{i}_1$ due to $\text{JS}(i_1, \hat{i}_1) = 1$. $i_2$ is aligned with "NULL". $i_4$ is aligned with $\hat{i}_1$ due to $\text{JS}(i_4, \hat{i}_1)=0.5$ and $\text{JS}(i_4, h)=0.4$. (3) We assign rewards based on the alignment results. Note that $i_3$ ($s_3$) is a redundant step. $r_{1}=1$, $r_{2}=-1$, $r_{3}=-0.5$, and $r_{4}=-1$. The reward for each state originates from the tree structure rather than the chained trajectory. Therefore, the return of each state should also follow the tree structure (or graph structure in reasoning graphs) rather than the chained trajectory.} 
    \label{fig:reward_process}
\end{figure*}

\begin{figure*}[]
    \centering
    \includegraphics[width=1\linewidth]{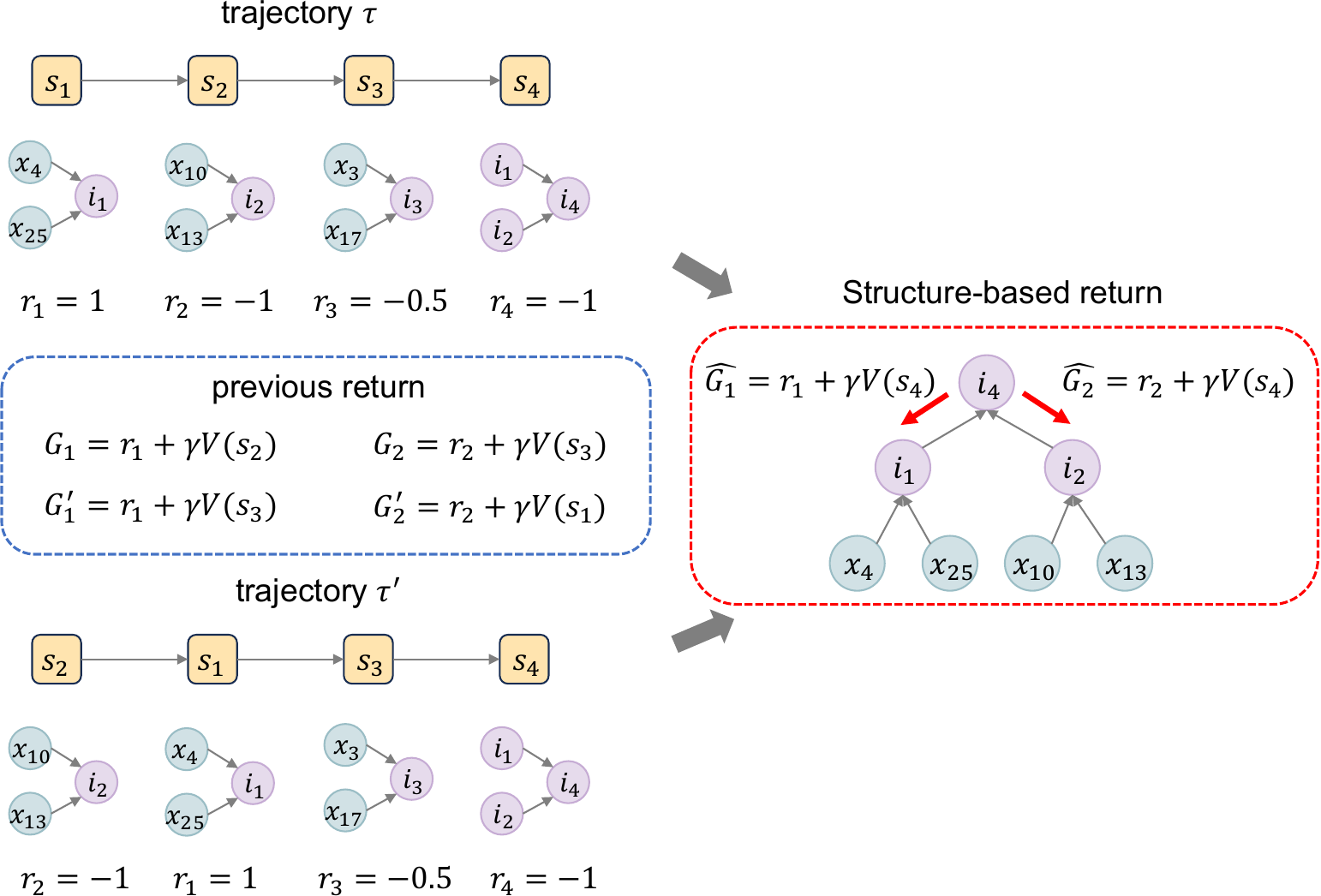}
    \caption{An illustration of the equivalent trajectory and the definition of return. As the reasoning steps of $s_1$, $s_2$, and $s_3$ are mutually independent, the execution order among these steps can be arbitrary. Thus, $\tau$ and $\tau^{\prime}$ are equivalent trajectories because they can be converted to the same entailment tree~\citep{dalvi-etal-2021-explaining}.
    As shown in \textcolor{blue}{blue box}, previous work~\citep{liu-etal-2022-rlet} defines the return (a.k.a cumulative reward) in a chained trajectory and would assign different returns to $s_1$ and $s_2$ in these equivalent trajectories. In contrast, as shown in \textcolor{red}{red box}, our structure-based return is defined based on the tree or graph structure inherent in structured reasoning, which is the same source of rewards. Our structure-based return will consistently allocate stable returns to equivalent trajectories, thereby promoting training stability and convergence. Furthermore, maintaining consistency between the sources of rewards and returns can significantly enhance the effectiveness of the policy.}
    \label{fig:equivalent_trajectory}
\end{figure*}

Given a hypothesis $h$ and initial facts $X$, we first obtain the trajectory through the reasoning process, as shown in Figure~\ref{fig:case_study}.
The state update follows the Markov decision process~\citep{bellman1957markovian}, meaning the current state only depends on the previous state.
Figure~\ref{fig:case_study} is an erroneous reasoning example to better illustrate the following steps.
Then, we convert the trajectory $\tau$ into an entailment tree $T_{\text{pred}}$ and align it with $T_{\text{gold}}$ to assign reward for each intermediate conclusion (as presented in Figure~\ref{fig:reward_process}).
Furthermore, Figure~\ref{fig:equivalent_trajectory} elucidates the issue of equivalent trajectories, and previous work can not accurately describe the logical relationship between different states in entailment trees.

\section{Prompts for GPT-4}
\label{sec:appendix_GPT4}
\begin{figure*}[t]
    \centering
    \includegraphics[width=1\linewidth]{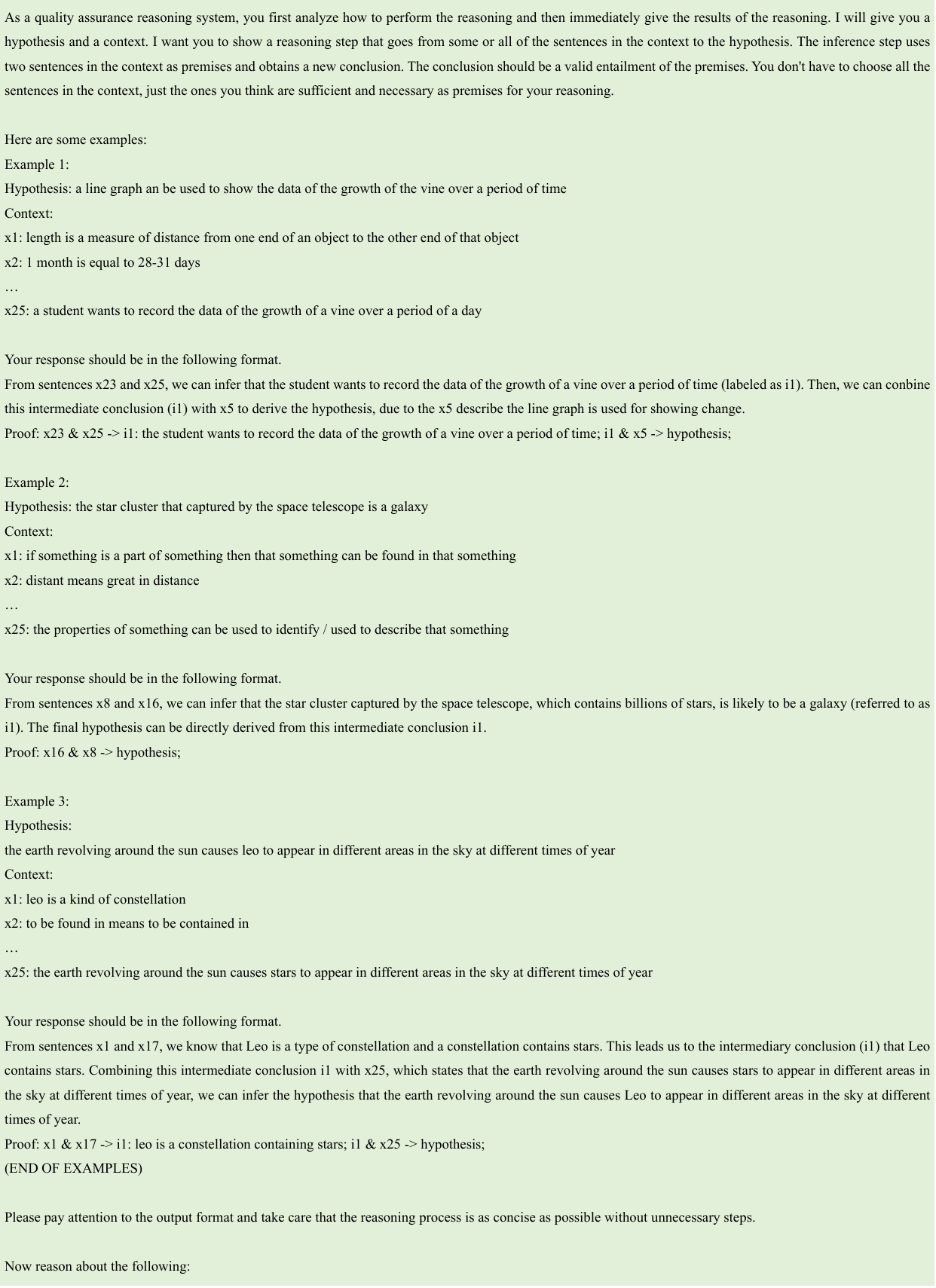}
    \caption{A Chain-of-Thought prompt for GPT-4.}
    \label{fig:prompt_cot}
\end{figure*}
\begin{figure*}[t]
    \centering
    \includegraphics[width=1\linewidth]{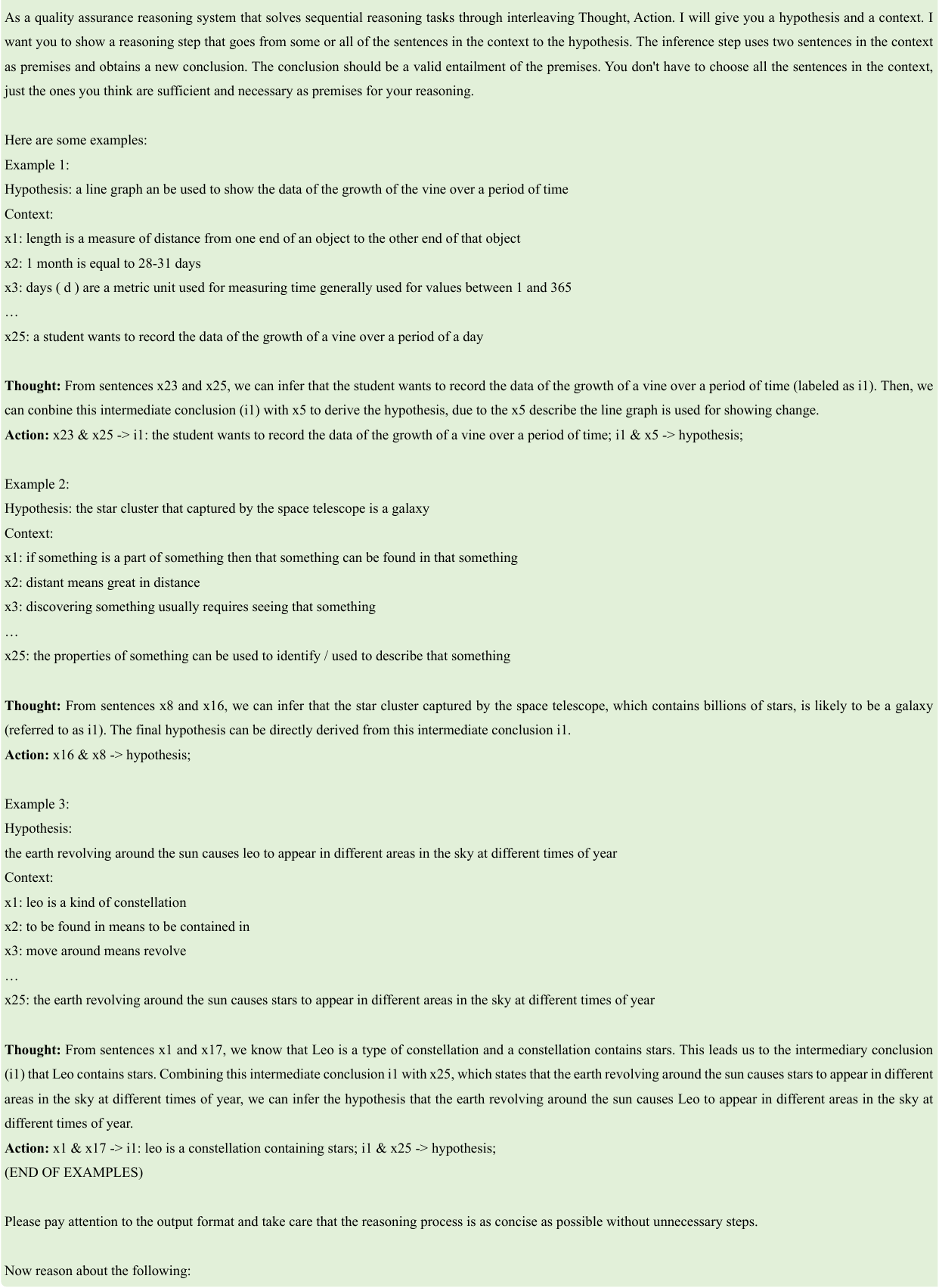}
    \caption{A ReAct prompt for GPT-4. "Thought" and "Action" query GPT-4 in two rounds.}
    \label{fig:prompt_react}
\end{figure*}
\begin{figure*}[t]
    \centering
    \includegraphics[width=1\linewidth]{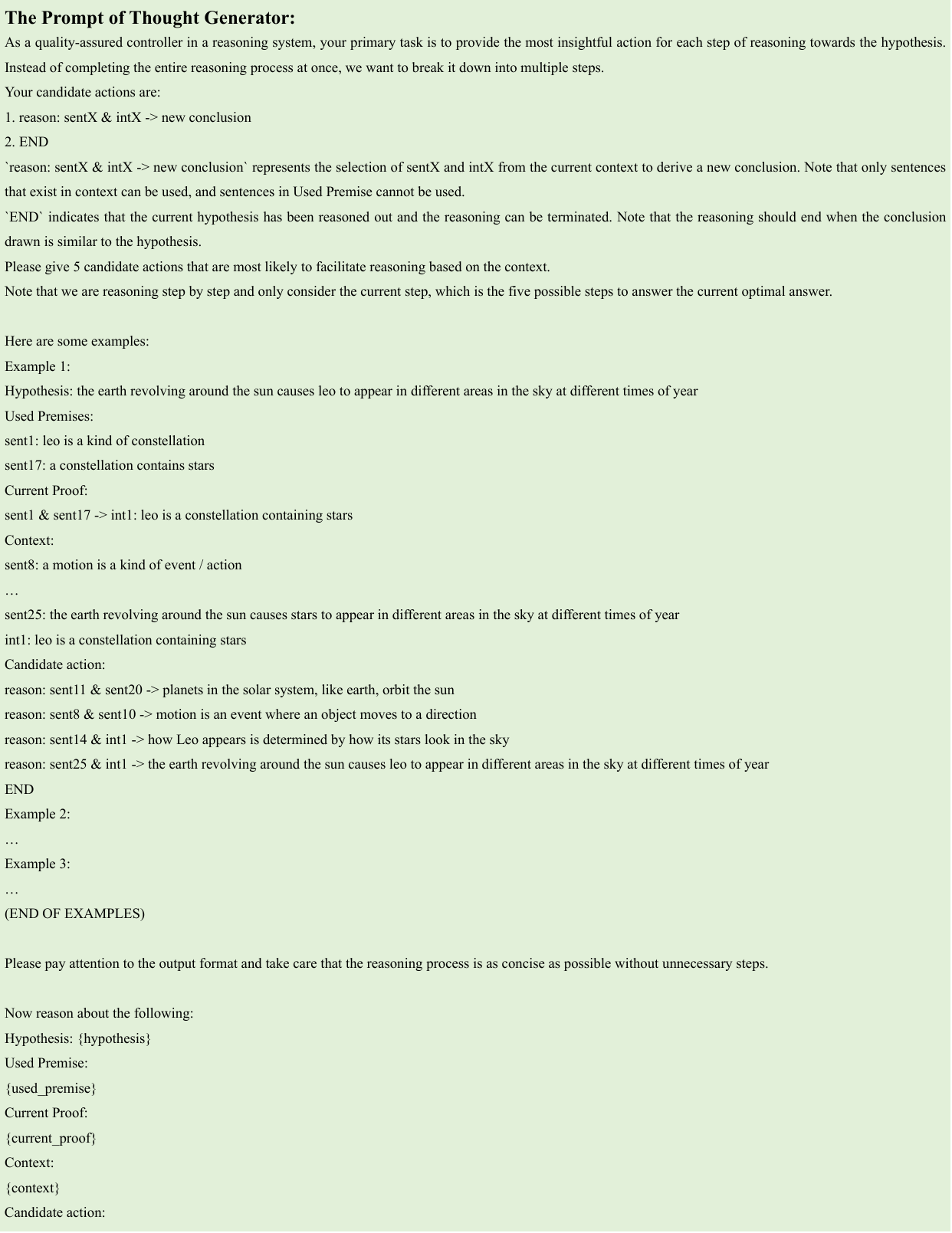}
    \caption{A Tree of Thought prompt (\textbf{Thought Generator}) for GPT-4.}
    \label{fig:tot_thought_generator}
\end{figure*}
\begin{figure*}[t]
    \centering
    \includegraphics[width=1\linewidth]{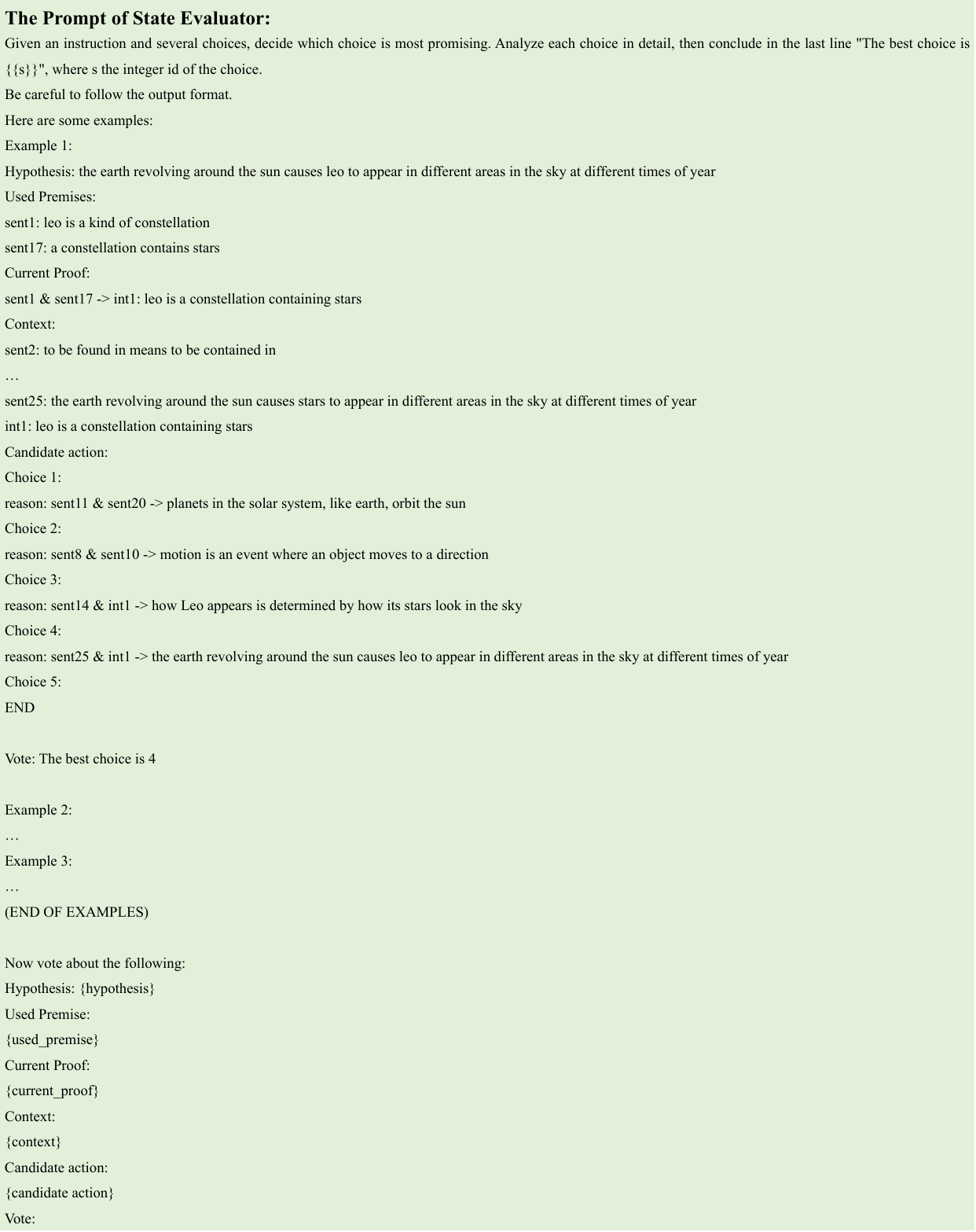}
    \caption{A Tree of Thought prompt (\textbf{State Evaluator}) for GPT-4.}
    \label{fig:tot_state_evaluator}
\end{figure*}

Figures~\ref{fig:prompt_cot} and~\ref{fig:prompt_react} show the Chain-of-Thought (CoT)~\citep{NEURIPS2022_9d560961} and ReAct~\citep{DBLP:conf/iclr/YaoZYDSN023} prompts for GPT-4, and figures~\ref{fig:tot_thought_generator} and~\ref{fig:tot_state_evaluator} show the prompts of thought generator and state evaluator in Tree of Thought (ToT)~\citep{yao2023tree}, respectively.
We provide a detailed introduction to the task definition and guide the model to respond in the required format.
We randomly selected three examples for in-context learning.
For a fair comparison, we use the same examples for CoT and ReAct, attributing similar thoughts to them.
ReAct divides the dialogue into two rounds, "Thought" and "Action", to query GPT-4.
For ToT, following~\citep{yao2023tree}, we generate candidate actions using a thought generator and subsequently select and execute the optimal action through a state evaluator.
Due to its exceptional reasoning capabilities and self-evaluation strategy, ToT achieves superior results compared to CoT and ReAct, as shown in Table~\ref{tab:entailmentbank}.
However, ToT requires higher costs in comparison to CoT and ReAct.

\section{Error Analysis}
\label{sec:appendix_error_analysis}
We conduct a comprehensive error analysis on Task2 and Task3 of EntailmentBank.

\subsection{Error Analysis of Task2}
\label{sec:appendix_error_analysis_task2}
We randomly sample 50 entailment trees where \modelname made incorrect reasoning. We find the following four types of errors.

\textbf{(1) Reasoning Step Error (62\%).}
This is the main source of errors and predominantly depends on whether the policy can select the correct premise.
We observe that a small portion of the errors (accounting for 12.9\% of this error type) use all the gold leaves, but have errors in the combination order.
Compared to other reasoning step errors, the policy identified the correct premise.
For example, the gold steps are "$x_{24} \; \& \; x_{5} \rightarrow i_1$; $i_1 \; \& \; x_{23} \rightarrow h$" and the error steps predicted by \modelname are "$x_{23} \; \& \; x_{5} \rightarrow i_1$; $i_1 \; \& \; x_{24} \rightarrow i_2$".

\textbf{(2) Early Termination Error (18\%).}
We observe that the reasoning process may terminate prematurely and the existing entailment steps are all correct.
On one hand, T5-large outputs ``End'' prematurely, unlike GPT-4 which can accurately judge when to stop.
On the other hand, the entailment module might erroneously infer a hypothesis, leading to premature termination.

\textbf{(3) Intermediate Conclusion Error (10\%).}
This error type is different from the above error (where the entailment module prematurely infers a hypothesis).
Intermediate conclusion error denotes errors triggered by incorrect entailment in the intermediate conclusions, despite having correct leaves and steps.
For a fair comparison, we used the entailment module that has already been trained in previous work~\cite{hong-etal-2023-faithful}. It is noted that the reasoning part, which is the focus of our paper, is completely correct in this type of error, and this type of error can be mitigated by training a better entailment module. 

\textbf{(4) Imperfect Evaluation (10\%).}
We discover that some trees deemed as invalid are valid in fact, indicating that current automated metrics underestimate the validity of the trees.
The most common reason is that there are multiple valid ways to construct an entailment tree.
For example, consider the structure of a gold tree:
"$x_1 \; \& \; x_2 \; \& \; x_3 \rightarrow h$" may be predicted as: "$x_1 \; \& \; x_2 \rightarrow i_1$; $i_1\; \& \; x_3 \rightarrow i_2$".

\subsection{Error Analysis of Task3}
Task 3 requires retrieving an initial set of facts $X$ from the corpus.
Therefore, in addition to the errors in Task 2 described above, we found that Task 3 has its own unique set of errors.

\textbf{(1) Missing Gold Leaves Error (58\%).}
Missing gold leaves error refers to the case where the gold leaves are not included in the facts $X$ retrieved from the corpus.
This case will inevitably lead to an error in the predicted entailment tree, regardless of how powerful the policy is.
The bottleneck of this error lies in the retrieval model.
For a fair comparison, we directly use the retrieval model provided in previous work~\citep{hong-etal-2023-faithful}.

\textbf{(2) Reasoning Errors (42\%).}
The four error types described in \ref{sec:appendix_error_analysis_task2} account for 42\% in Task3.

We also discovered that the reasoning graph contains similar error types as found in entailment trees, as they both belong to structured reasoning.

\end{document}